\documentclass[sigconf]{acmart}
\usepackage{multirow}  
\usepackage{setspace}  
\AtBeginDocument{%
  }

\copyrightyear{2025}
\acmYear{2025}
\setcopyright{cc}
\setcctype{by-nc-sa}
\acmConference[SIGIR '25]{Proceedings of the 48th International ACM SIGIR Conference on Research and Development in Information Retrieval}{July 13--18, 2025}{Padua, Italy}
\acmBooktitle{Proceedings of the 48th International ACM SIGIR Conference on Research and Development in Information Retrieval (SIGIR '25), July 13--18, 2025, Padua, Italy}\acmDOI{10.1145/3726302.3729938}
\acmISBN{979-8-4007-1592-1/2025/07}

\begin{document}

\title{CSE-SFP: Enabling Unsupervised Sentence Representation Learning via a Single Forward Pass}

\author{Bowen Zhang}
\email{zbw23@mails.tsinghua.edu.cn}
\orcid{0009-0001-2550-6468}
\affiliation{%
  \institution{Tsinghua University}
  \city{Beijing}
  \country{China}}

\author{Zixin Song}
\email{songzx24@mails.tsinghua.edu.cn}
\orcid{0009-0008-9111-3866}
\affiliation{%
  \institution{Tsinghua University}
  \city{Beijing}
  \country{China}}

\author{Chunping Li}
\authornote{Chunping Li is the corresponding author.}
\email{cli@tsinghua.edu.cn}
\orcid{0000-0002-4521-0875}
\affiliation{%
  \institution{Tsinghua University}
  \city{Beijing}
  \country{China}}

\renewcommand{\shortauthors}{Bowen et al.}

\begin{abstract}
As a fundamental task in Information Retrieval and Computational Linguistics, sentence representation has profound implications for a wide range of practical applications such as text clustering, content analysis, question-answering systems, and web search. Recent advances in pre-trained language models (PLMs) have driven remarkable progress in this field, particularly through unsupervised embedding derivation methods centered on discriminative PLMs like BERT. However, due to time and computational constraints, few efforts have attempted to integrate unsupervised sentence representation with generative PLMs, which typically possess much larger parameter sizes. Given that state-of-the-art models in both academia and industry are predominantly based on generative architectures, there is a pressing need for an efficient unsupervised text representation framework tailored to decoder-only PLMs. To address this concern, we propose CSE-SFP, an innovative method that exploits the structural characteristics of generative models. Compared to existing strategies, CSE-SFP requires only a single forward pass to perform effective unsupervised contrastive learning. Rigorous experimentation demonstrates that CSE-SFP not only produces higher-quality embeddings but also significantly reduces both training time and memory consumption. Furthermore, we introduce two ratio metrics that jointly assess alignment and uniformity, thereby providing a more robust means for evaluating the semantic spatial properties of encoding models. Our code and checkpoints are available at \url{https://github.com/ZBWpro/CSE-SFP}.
\end{abstract}




\maketitle

\section{Introduction}

Sentence representation learning aims to map natural language inputs into fixed-length numerical vectors, commonly referred to as text embeddings, which can be processed by computational systems and neural networks. These encodings are pivotal for Information Retrieval (IR), as they capture the semantic essence of original texts while exhibiting strong transferability. As a result, sentence representations underpin diverse real-world applications, including search engines, recommendation systems, dialogue platforms, and retrieval-augmented generation (RAG) \cite{BERT4Rec-CIKM-2019, RAG-Survey-2024}.

Since the introduction of seminal works like Sentence-BERT \cite{SBERT-EMNLP-2019} and SimCSE \cite{SimCSE-EMNLP-2021}, substantial strides have been made in sentence representation schemes based on discriminative PLMs, exemplified by BERT \cite{BERT-NAACL-2019} and RoBERTa \cite{RoBERTa-2019}. Among these, unsupervised contrastive learning methods, where models are trained on corpora consisting solely of individual sentences, have become a focal point in natural language processing (NLP) and IR research \cite{SR-Survey-EACL-2024}, giving rise to a considerable body of studies \cite{CARDS-SIGIR-2022, PromptBERT-EMNLP-2022, DiffCSE-NAACL-2022, SNCSE-ICIC-2023, CoT-BERT-ICANN-2024}. 

With the rapid development of large language models (LLMs), cutting-edge approaches such as PromptEOL \cite{PromptEOL-EMNLP-2024}, DeeLM \cite{DeeLM-2023}, and Pcc-tuning \cite{Pcc-tuning-EMNLP-2024} have opted to utilize generative PLMs with larger parameter scales (e.g., 7B) for \textbf{supervised} sentence representation, yielding impressive results. In contrast, only a limited number of research has explored the use of these models for \textbf{unsupervised} sentence embedding derivation \cite{LLM2Vec-COLM-2024}. The primary reason for this gap probably stems from the fact that, compared to supervised corpora rich in annotated information, unsupervised data offer far less prior knowledge and much fewer semantic signals. Consequently, larger text volumes are needed, which drastically increases training costs (see Table~\ref{tab:sup_vs_unsup}). For instance, the supervised dataset adopted by SimCSE contains 275,601 samples, whereas its unsupervised counterpart encompasses as many as 1,000,000 entries \cite{SimCSE-EMNLP-2021}. Considering that a 7B-scale PLM has over 60 times the parameter count of BERT\(_\text{base}\), coupled with the necessity of large batch sizes for contrastive learning to avoid model collapse \cite{STS-Reg-EMNLP-2024}, the computational overhead becomes prohibitively expensive.

\begin{table}[htbp]
\caption{Training time and GPU memory usage of $\rm{Mistral}_{\rm 7b}$ when fine-tuned with supervised and unsupervised datasets for contrastive learning. Our proposed CSE-SFP significantly improves both training and memory efficiency.}
\resizebox{1.0\linewidth}{!}{
    \renewcommand
    \arraystretch{1.3}
    \centering
    \begin{tabular}{cccc}
    \hline
    Methods & Samples & Training Time & Memory Usage \\
    \midrule
    Supervised SimCSE & 275,601 & 116.89 min & 92.67 GB\\
    Unsupervised SimCSE & 1,000,000 & 292.92 min & 85.82 GB\\
    CSE-SFP (Unsupervised) & 1,000,000 & \underline{189.68} min & \textbf{80.29} GB\\
    \hline
    \end{tabular}
}
\label{tab:sup_vs_unsup}
\end{table}

Given that high-quality supervised corpora are often scarce and costly to annotate in downstream tasks \cite{BERT-flow-EMNLP-2020}, unsupervised sentence representation methods that do not rely on labeled data hold great promise for both research and practical applications. To realize the potential of generative PLMs for unsupervised text representation, it is essential to mitigate the associated computational costs. 

Currently, mainstream unsupervised sentence embedding strategies generally employ contrastive learning to refine the model's semantic space \cite{Pcc-tuning-EMNLP-2024}. However, contrastive loss functions require embeddings of semantically similar content to form positive sample pairs. In unsupervised settings, positive examples are typically constructed through data augmentation techniques such as dropout, Gaussian noise, or truncation \cite{ConSERT-ACL-2021, SimCSE-EMNLP-2021, CLSEP-KBS-2023}. This means that the same piece of text \(x_i\) must be fed into the model twice, undergoing two separate forward passes to calculate its own encoding \(f(x_i)\) and that of its augmented version \(f(x_i)^+\). This duplication inevitably leads to huge memory consumption and training delays.

Unlike discriminative PLMs based on Transformer encoder architectures \cite{Transformer-NIPS-2017}, generative models are pre-trained with autoregressive language modeling and employ a unidirectional attention mechanism. That is, for any given position \(p\), the model cannot attend to tokens that follow it. This structural property motivates us to design a two-stage prompt to encapsulate the input sentence \( x_i \), where each stage incorporates a representation token dedicated to extracting embeddings. By doing so, we can leverage both the model's encoding and generative capabilities to obtain \(f(x_i)\) and \(f(x_i)^+\) simultaneously within a single forward pass.

Although both \( f(x_i) \) and \( f(x_i)^+ \) represent the same text \( x_i \), their vector compositions exhibit inherent discrepancies due to variations in guiding templates, embedding collection positions, and attention scopes. Thus, these two sets of embeddings are maximally differentiated while preserving semantic similarity, fulfilling the contrastive learning requirement for positive pairs to be semantically close yet distinct \cite{DiffAug-EMNLP-2022}.

\begin{figure}[htbp]
\centering
\includegraphics[width=\linewidth]{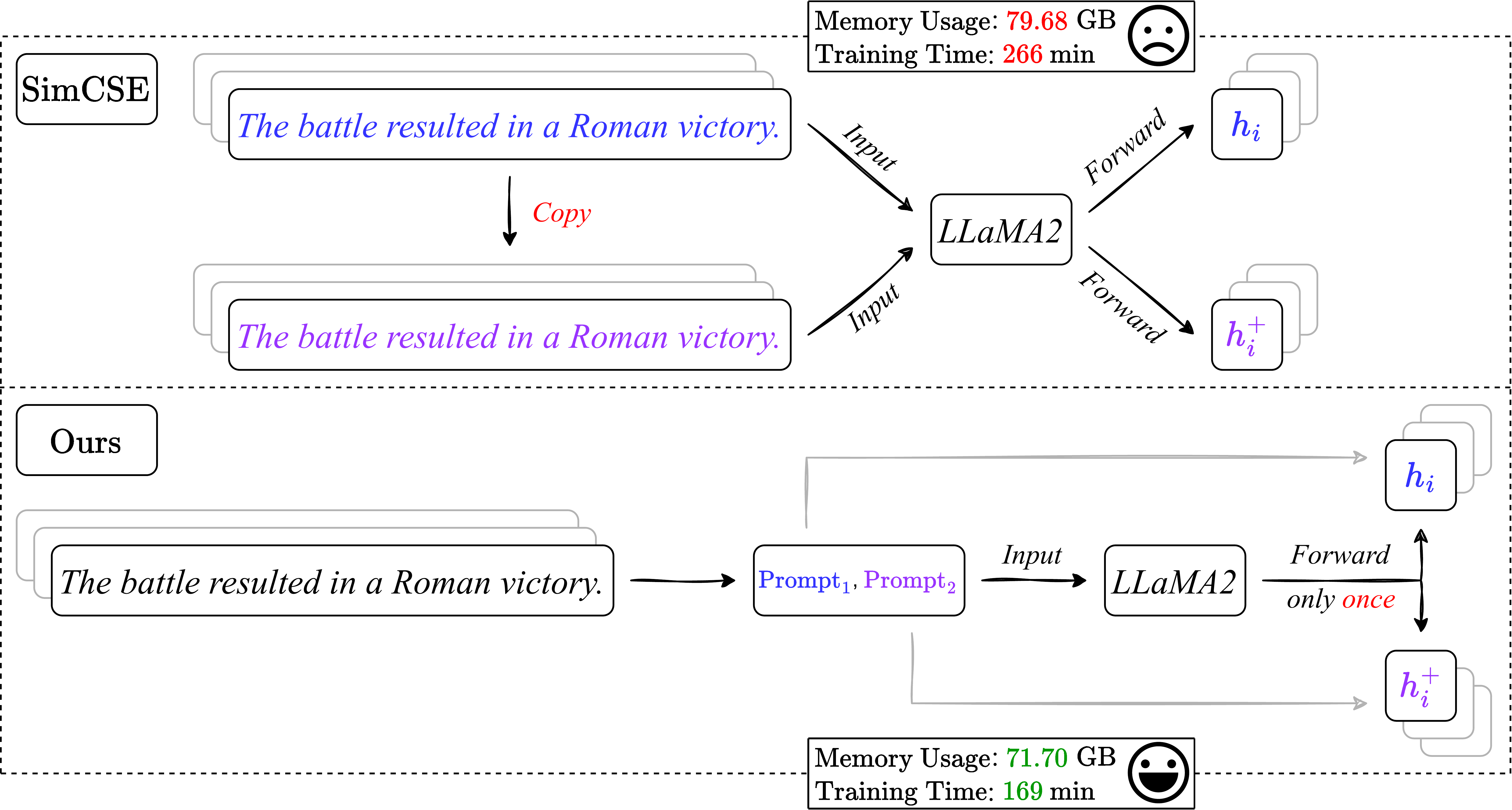}
\caption{Workflow comparison between traditional methods (e.g., SimCSE) and CSE-SFP. SimCSE generates positive samples via built-in dropout within the Transformer block, requiring an additional copy of the same text and performing two forward computations to acquire the anchor sentence embedding \( h_i \) and its positive counterpart \( h_i^+ \). In contrast, CSE-SFP concatenates two distinct manual templates, allowing both embeddings to be generated in a single forward pass.}
\Description{A diagram comparing the SimCSE and CSE-SFP workflows for contrastive learning. The CSE-SFP pipeline is more streamlined and efficient, involving only one forward pass compared to SimCSE's two.}
\label{fig:paradigm}
\end{figure}

Building on these insights, we propose CSE-SFP: an unsupervised \underline{C}ontrastive \underline{S}entence \underline{E}mbedding framework that requires only a \underline{S}ingle \underline{F}orward \underline{P}ass to facilitate effective contrastive training. Figure~\ref{fig:paradigm} illustrates the differences between our method and conventional unsupervised sentence representation approaches, with more detailed comparisons and discussions provided in subsequent sections. The main contributions of this paper are as follows:
\begin{itemize}
\item We perform a thorough evaluation of existing generative PLM-based sentence embedding methods from multiple perspectives, including representation quality, memory usage, and training time, establishing important baselines for future research. 

\item We introduce CSE-SFP, a streamlined unsupervised sentence representation framework. Distinct from existing contrastive learning methods, CSE-SFP needs only a single forward propagation per text to simultaneously generate the anchor embedding and its positive counterpart, greatly simplifying the contrastive learning process. Experimental results across various backbone models demonstrate that CSE-SFP not only serves as a versatile data augmentation strategy but also outperforms prevalent dropout-based techniques for positive sample construction.

\item We extensively validate the superiority of CSE-SFP in terms of performance, efficiency, and resource utilization across seven internationally recognized Semantic Textual Similarity (STS) benchmarks and eight IR tasks. To further elucidate the underlying mechanisms of our method, we conduct a series of theoretical analyses, revealing that CSE-SFP significantly enhances the representational capacity of text embeddings. Additionally, drawing on the concepts of alignment and uniformity, we propose two novel ratio-based metrics for a more comprehensive assessment of PLMs' semantic space.
\end{itemize}

\section{Background and Related Work}

\subsection{Task Definition}

This paper focuses on general-purpose sentence representation. For any given natural language text \( x \), the goal is to design an appropriate mapping function \( f \) that transforms \( x \) into a \( d \)-dimensional vector encoding \( f(x) \). To meet the efficiency requirements of large-scale information retrieval, the distance between sentence embeddings should accurately reflect the semantic relevance of their corresponding texts. Specifically, if the semantic similarity between \( x_1 \) and \( x_2 \) is higher than that between \( x_3 \) and \( x_4 \), a well-performing mapping \( f \) should satisfy \( \text{dis}(f(x_1), f(x_2)) < \text{dis}(f(x_3), f(x_4)) \). Typically, the distance metric "\(\text{dis}\)" is chosen to be a simple and rapidly computable measure like cosine similarity \cite{SBERT-EMNLP-2019}. In this case, we would have \( \cos(f(x_1), f(x_2)) > \cos(f(x_3), f(x_4)) \).

\subsection{Contrastive Learning for Sentence Embedding}

There exists a strong connection between the objective outlined above and contrastive learning. Given a batch of texts \(\{x_i\}_{i=1}^N\), contrastive loss functions, such as InfoNCE Loss \cite{InfoNCE-2018}, calculate the similarity between each sample \( x_i \) and its positive example in the numerator of a cross-entropy function, while aggregating the similarities between \( x_i \) and other texts within the same batch in the denominator. The mathematical expression for InfoNCE Loss is given by Equation~\ref{eq:infonce}, where \(\tau\) denotes a temperature hyperparameter. This formulation aims to maximize the probability that \( f(x_i) \) is classified into the same category as \( f(x_i)^+ \). In unsupervised text representation tasks, the positive example for \( x_i \) is unknown and must be constructed manually.
\begin{equation}
    \label{eq:infonce}
    \mathcal{\ell}_{i} = - \log \frac{e^{\cos(f(x_i), f(x_i)^+) / \tau}}{\sum_{j=1}^N e^{\cos(f(x_i), f(x_j)^+) / \tau}}
\end{equation}

Intuitively, contrastive learning can be viewed as a form of clustering at the sample level, which encourages the representations of different texts to be as distinct as possible. Previous research has shown that contrastive learning can significantly enhance the uniformity of embeddings while maintaining the alignment of the PLM's semantic space \cite{SimCSE-EMNLP-2021}, thus making the embeddings distribution more suitable for metrics such as cosine similarity. In this context, leveraging contrastive learning to improve representation quality has become a consensus within the AI community \cite{CLSEP-KBS-2023}. Therefore, the construction of positive samples \( f(x_i)^+ \) is particularly critical, as it directly influences the effectiveness of contrastive learning.

\subsection{Constructing Positive Examples}

Over the past few years, unsupervised embedding derivation methods for BERT-style discriminative PLMs have dominated the research landscape in sentence representation \cite{SR-Survey-EACL-2024}. A key challenge in this domain is how to create positive examples that are semantically close to the input text without relying on any annotated information. Researchers have devised various solutions to this problem. Among them, ConSERT \cite{ConSERT-ACL-2021} leverages four strategies, including token shuffling and adversarial attacks, to construct positive samples. Subsequently, SimCSE \cite{SimCSE-EMNLP-2021} discovered that standard dropout can generate positive embeddings superior to those produced by discrete data augmentation strategies such as word deletion and synonym replacement. Building upon this, ESimCSE \cite{ESimCSE-COLING-2022} further improves the approach by repeating words in the input text, thereby overcoming the limitation in SimCSE where positive samples are always the same length as the original sentence. CARDS \cite{CARDS-SIGIR-2022}, on the other hand, randomly flips the first letter of words to alleviate the model's bias towards case sensitivity.
 
Despite their success, all of these methods require two forward passes to obtain both \( f(x_i) \) and \( f(x_i)^+ \). As model sizes and training datasets continue to grow, the time and memory costs of this process become increasingly burdensome. Moreover, the dropout mechanism, which forms the core of these strategies \cite{PromptBERT-EMNLP-2022}, is not universally available in generative PLMs (e.g., LLaMA2 \cite{LLaMA2-2023}), potentially leading to inconsistent benefits when transferring these techniques to LLMs.

\section{Methodology}

This section introduces CSE-SFP, an innovative unsupervised sentence representation framework. First, in subsection~\ref{sec:motivation}, we explain the design principles of our approach by integrating the structural characteristics of generative PLMs and the implementation of autoregressive language modeling. Then, in subsection~\ref{sec:cse-sfp}, we present the overall architecture of CSE-SFP, along with its training and inference workflows.

\subsection{Motivation}
\label{sec:motivation}

\textbf{Observation 1: LLMs Possess Both Encoding and Generative Capabilities}

As highlighted by GRIT \cite{GRIT-2024}, all text-oriented language problems can be simplified into two broad categories: embedding and generation. Leveraging their vast parameter scales and abundant pre-training corpora, generative PLMs have demonstrated exceptional performance across diverse IR and NLP tasks since their inception \cite{GPT3-2020-NIPS, Mistral-2023}. This success indicates that modern LLMs are equipped with robust semantic understanding and text continuation abilities.

This assertion finds strong support when examining the model structure and pre-training objectives of LLMs. Consider an input sequence \( T = [t_1, t_2, \ldots, t_n] \), where each \( t_i \) is a token resulting from word segmentation. Firstly, the model maps each token \( t_i \) into a \( d \)-dimensional dense vector \( x_i \) via an embedding layer and adds positional encodings to form the initial word embedding matrix \( X \in \mathbb{R}^{n \times d} \). At this stage, each row \( x_i \in \mathbb{R}^d \) in \( X \) remains relatively independent. However, tokens within a natural language text are inherently interconnected. The same word can exhibit different semantic nuances depending on its context. To model these inter-token dependencies, Transformer \cite{Transformer-NIPS-2017} employs an attention mechanism, utilizing three learnable matrices \( W_Q, W_K, W_V \in \mathbb{R}^{d \times d_k} \) to project \( X \) into query \( Q = XW_Q \), key \( K = XW_K \), and value \( V = XW_V \). Subsequently, attention scores are computed to yield contextually weighted token representations:
\begin{equation}
    \label{eq:attn}
    \text{Attention}(\mathbf{Q}, \mathbf{K}, \mathbf{V}) = \text{Softmax}\left(\frac{\mathbf{Q} \mathbf{K}^\top}{\sqrt{d_k}}\right) \mathbf{V}
\end{equation}

On the decoder side, a causal mask is applied to the attention distribution, ensuring that \( x_i \) only depends on preceding tokens \( x_1, \ldots, x_i \), thereby preventing information leakage. Additionally, the process described above pertains to a single attention head. The outputs from different attention heads are typically concatenated and fused through a series of linear layers, coupled with residual connections \cite{ResNet-CVPR-2016} and layer normalization, to produce the input embedding matrix \( X^l \in \mathbb{R}^{n \times d} \) for the next Transformer block, where \( l \) signifies the layer index.

Notably, although \( X^l \) shares the same dimensions as \( X \), the information it contains has evolved. Each row vector \( x_i^l \) in \( X^l \) no longer merely represents the superficial meaning of the token \( t_i \) itself, but rather its semantic role within the entire sequence. In other words, the interactive effect of attention enables each token to aggregate information from other tokens according to their relevance, thus endowing individual words with sentence-level expressions. Consequently, after processing by multiple Transformer layers, the entire input sequence \( T \) is encoded. Therefore, LLMs, with their stacked Transformer architecture, inherently possess potent encoding capabilities.

The generative power of LLMs, on the other hand, arises from their pre-training task: autoregressive language modeling. Given an input sequence \( T = [t_1, t_2, \ldots, t_n] \), the model's prediction target can be viewed as a shifted version of \( T \), denoted \( T' = [t_2, \ldots, t_{n+1}] \). For any subsequence \( t_1, t_2, \ldots, t_{i-1} \) of \( T \), the PLM calculates the probability of sampling the next token \( t_i \) based on the state vector \( x_{i-1}^L \) corresponding to $t_{i-1}$ from the final hidden layer $L$, in conjunction with an output projection head $W_\text{out}$:
\begin{equation}
    \label{eq:sample}
    \text{P}(t_i \mid t_1, t_2, \ldots, t_{i-1}) = \text{Softmax}\left(x_{i-1}^L W_\text{out} \right)
\end{equation}

Under this training paradigm, the final word embedding \( x_{i-1}^L \) not only captures the contextual semantics of \( t_{i-1} \), but also carries indicative information about the upcoming token \( t_i \). The latter aspect is a key manifestation of the model's generative prowess. This dual nature of LLMs suggests a novel direction: for a given text segment, if we can mobilize different aspects of the model to derive two separate embeddings, they could potentially form effective positive sample pairs for contrastive learning. 

\textbf{Observation 2: The Attention Mechanism in Generative PLMs is Unidirectional}

As mentioned earlier, to maintain the autoregressive property for language generation, the self-attention computation in the Transformer decoder incorporates a causal mask. Specifically, extending Equation~\ref{eq:attn}, an upper-triangular mask matrix \( M \in \mathbb{R}^{n \times n} \) is introduced and added to the scaled dot-product scores before applying the Softmax function:
\begin{equation}
    \text{Attention}(\mathbf{Q}, \mathbf{K}, \mathbf{V}) = \text{Softmax}\left(\frac{\mathbf{Q} \mathbf{K}^\top}{\sqrt{d_k}} + \mathbf{M}\right) \mathbf{V}
\end{equation}

Here, elements above the main diagonal of \( M \) are set to negative infinity, while all other elements are zero. This masking guarantees that a token at position \( i \) cannot observe tokens appear later in the sequence. Therefore, if a template comprising two parts is fed into the model, the word embeddings within the "Prefix" will not be influenced by the "Suffix":
\begin{equation}
    \text{Template} = \text{Concat}(\text{Prefix}, \text{Suffix})
\end{equation}

We can exploit this property for data augmentation by designing both the "Prefix" and "Suffix" as prompts that guide the PLM to represent the input sentence. From the perspective of the "Prefix", the "Suffix" functions as an independent statement, making the entire process approximate the use of distinct manual templates. Furthermore, we ensure through differential settings that the "Suffix" does not produce embeddings identical to those of the "Prefix". These details will be elaborated upon in the subsequent section.

\subsection{CSE-SFP}
\label{sec:cse-sfp}

\begin{figure*}[ht]
\centering
\includegraphics[width=1.0\linewidth]{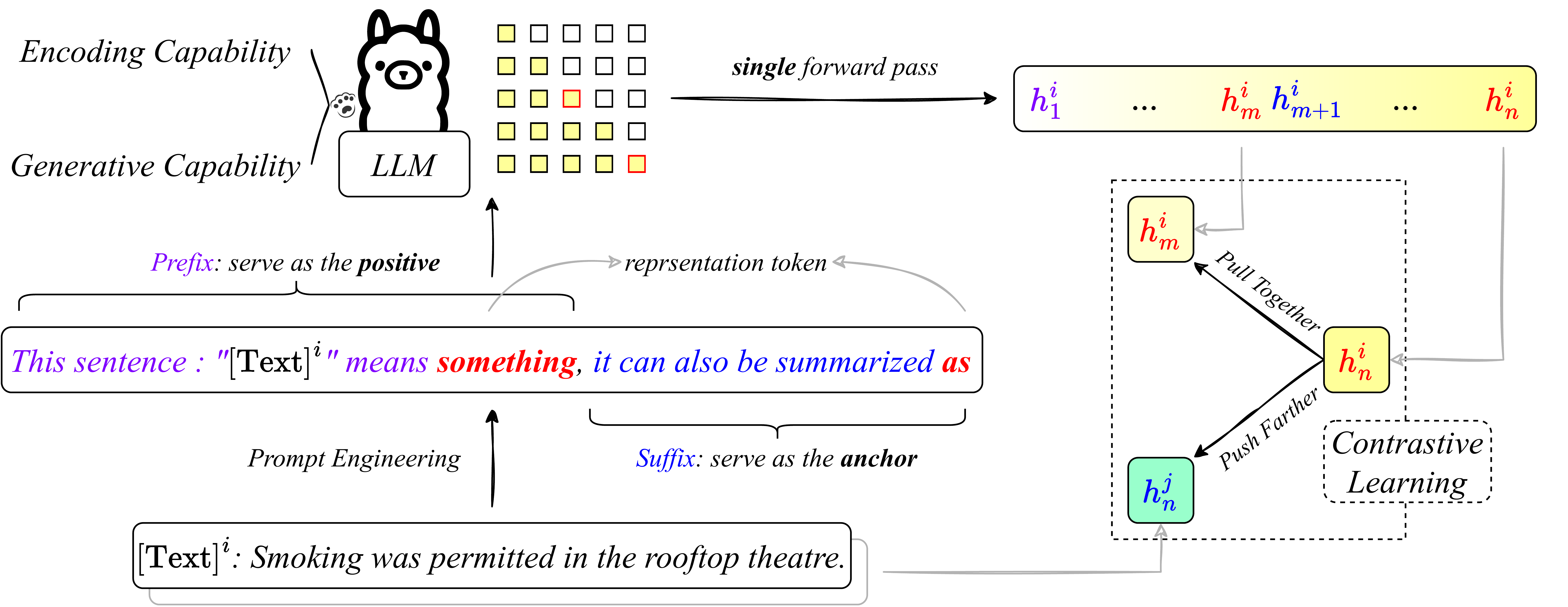}
\caption{The overall architecture of CSE-SFP. By taking full advantage of LLMs' structural as well as functional characteristics, we obtain \( h_m^i \) and \( h_n^i \) for constructing positive sample pairs in contrastive learning with just a single forward pass. Moreover, both the prefix and suffix of CSE-SFP are flexible, allowing for customization based on different PLMs and downstream tasks. Here, we exemplify the assembly of a two-stage prompt using PromptSTH and PromptSUM, as proposed by PretCoTandKE \cite{PretCoTandKE-ICIC-2024}.}
\Description{CSE-SFP's architecture.}
\label{fig:workflow}
\end{figure*}

Building upon these insights, we propose CSE-SFP, a novel text representation method tailored for generative PLMs. Figure~\ref{fig:workflow} depicts the overall architecture of CSE-SFP. For any input sentence \(\text{[Text]}^i\), we encapsulate it with a two-stage prompt, where each stage incorporates a representation token \( \text{Rep} \) to facilitate embedding extraction:
\begin{equation}
    \label{eq:template}
    \text{Template} = \text{Pre}_1 \ldots\textcolor{red}{\text{[Text]}^i}\ldots\text{Pre}_\text{m}\textcolor{purple}{\text{Rep}_1},\text{Suf}_1 \ldots \text{Suf}_\text{n} \textcolor{blue}{\text{Rep}_2}
\end{equation}

In this template, \(\text{Pre}_{1:m}\) forms the prefix portion, guiding the model to focus the semantics of \(\text{[Text]}^i\) onto the representation token \(\text{Rep}_1\). Conversely, \( \text{Suf}_{1:n} \) constitutes the suffix, inducing the PLM to generate vocabulary at the end of the sequence that summarizes the overall meaning of \(\text{[Text]}^i\). As shown in Equation~\ref{eq:sample}, the model predicts the next token based on the output vector of the last position. Therefore, the encoding of \(\text{Rep}_2\) inherently contains indicative information about the target word, making it a suitable representation of the original sentence.

This design enables the simultaneous acquisition of two sentence representations in a single forward pass, both of which are sufficiently diverse to support effective contrastive learning. Specifically, since \(\text{Rep}_1\) resides in the middle of the prompt, the model primarily relies on its encoding capabilities to compute the embedding. In contrast, \(\text{Rep}_2\) is not only located at the end of the template, but the suffix itself does not form a complete sentence. As a result, the output vector for \( \text{Rep}_2 \) is heavily dependent on the model's generative abilities. Furthermore, the positional encodings and attention scopes for \( \text{Rep}_1 \) and \( \text{Rep}_2 \) are also distinct. \(\text{Rep}_1\) interacts exclusively with the prefix of the template, ensuring that its embedding remains unaffected by the suffix. Although \(\text{Rep}_2\) can observe \(\text{Rep}_1\), it is guided by the instruction to produce an expression that is distinguishable from \( \text{Rep}_1 \).


It is important to note that CSE-SFP, as a general text representation framework, can adeptly accommodate various types of templates and is not confined to specific prompt configurations. Currently, there are three commonly adopted templates for deriving sentence embeddings from generative PLMs in the academic literature: PromptEOL \cite{PromptEOL-EMNLP-2024}, PromptSUM, and PromptSTH \cite{PretCoTandKE-ICIC-2024}, as detailed in Table~\ref{tab:templates}. It can be seen that PromptEOL and PromptSUM primarily leverage the model's generative capabilities, whereas PromptSTH tends to utilize the PLM's encoding abilities. Previous studies have found that, in supervised settings, the final results of these three approaches are quite comparable \cite{PretCoTandKE-ICIC-2024}. In this paper, we evaluate the performance of these templates under unsupervised settings, thereby establishing essential baselines for future research. In Figure~\ref{fig:workflow}, we illustrate how PromptSTH and PromptSUM can be integrated into CSE-SFP, with other combinations following a similar pattern.
\begin{table}[htbp]
\caption{Three mainstream sentence representation templates, where the red-highlighted parts indicate the position from which the model extracts embeddings.}
\centering
\begin{spacing}{1.3}
\begin{tabular}{c}
\toprule
{\bf PromptEOL}  \\
This sentence : "[Text]" means in one word:\textcolor{red}{"} \\
\midrule 
{\bf PromptSUM} \\
This sentence : "[Text]" can be summarized \textcolor{red}{as}\\
\midrule 
{\bf PromptSTH} \\
This sentence : "[Text]" means \textcolor{red}{something} \\
\bottomrule 
\end{tabular}
\end{spacing}
\label{tab:templates}
\end{table}

Regarding the workflow, we employ the standard InfoNCE loss function, as described in Equation~\ref{eq:infonce}, during the training phase. The output vectors from \( \text{Rep}_1 \) and \( \text{Rep}_2 \) are designated as the positive instance embedding \( f(x_i)^+ \) and the anchor sentence embedding \( f(x_i) \), respectively. By doing so, we effectively circumvent the need to duplicate each input text and perform separate encodings for \( f(x_i) \) and \( f(x_i)^+ \). During the testing phase, we directly utilize the output vector of \(\text{Rep}_2\) as the final sentence representation. A potential enhancement involves combining \(\text{Rep}_1\) and \(\text{Rep}_2\) in a manner that achieves more comprehensive expressive power, which we plan to explore in future work.

\section{Experiments}

This section provides empirical validation for our proposed CSE-SFP. First, in subsection~\ref{sec:implementation}, we outline the experimental setup of this study, including training procedures, evaluation benchmarks, and the selection of baselines. Following this, in subsections~\ref{sec:sts_performance} and \ref{sec:ir_performance}, we present the performance of CSE-SFP on Semantic Textual Similarity (STS) and Information Retrieval (IR) tasks, respectively. Finally, in subsection~\ref{sec:comparison}, we highlight the advantages of our method in terms of training time and memory consumption through comparative analysis.

\subsection{Implementation Details}
\label{sec:implementation}

In line with standard practices in unsupervised text representation research, we train the models on a corpus comprising one million randomly sampled sentences from English Wikipedia. This dataset was created by SimCSE and has been widely used for fine-tuning BERT \cite{SimCSE-EMNLP-2021, ESimCSE-COLING-2022, PromptBERT-EMNLP-2022, CoT-BERT-ICANN-2024}. To fully demonstrate the generality of our strategy, we evaluate CSE-SFP with four generative PLMs released at different times: OPT\(_\text{6.7b}\) \cite{OPT-2022}, LLaMA2\(_\text{7b}\) \cite{LLaMA2-2023}, Mistral\(_\text{7b}\) \cite{Mistral-2023}, and LLaMA3\(_\text{8b}\) \cite{LLaMA3-2024}. Given the substantial parameter sizes of these models, we adopt the same QLoRA \cite{QLoRA-NIPS-2024} configuration as PromptEOL \cite{PromptEOL-EMNLP-2024} and Pcc-tuning \cite{Pcc-tuning-EMNLP-2024} to mitigate computational overhead throughout all experiments.

In terms of evaluation benchmarks, STS tasks have long been regarded as the primary means of assessing sentence embeddings \cite{SBERT-EMNLP-2019, SimCSE-EMNLP-2021, Pcc-tuning-EMNLP-2024, SR-Survey-EACL-2024}. Therefore, we utilize the SentEval \cite{SentEval-LREC-2018} toolkit to test model performance across seven widely recognized STS datasets. Additionally, we select eight IR tasks from the recently introduced MTEB \cite{MTEB-EACL-2023} leaderboard to showcase CSE-SFP’s potential in practical applications.

As for baselines, we mainly compare our method against three leading contrastive learning approaches: PromptEOL \cite{PromptEOL-EMNLP-2024}, PromptSTH, and PromptSUM \cite{PretCoTandKE-ICIC-2024}. Previous studies have shown that these methods outperform directly transferring SimCSE to LLMs in both supervised fine-tuning and direct inference scenarios \cite{PromptEOL-EMNLP-2024, Pcc-tuning-EMNLP-2024, PretCoTandKE-ICIC-2024}. Notably, since the prefix and suffix of CSE-SFP's two-stage template are derived from these three methods in our experiments, the comparison between CSE-SFP and them also functions as an ablation study.

\subsection{Performance on Semantic Textual Similarity Tasks}
\label{sec:sts_performance}

Table~\ref{tab:sts_results} reports the Spearman correlation coefficients of various sentence representation methods on the seven STS tasks collected in SentEval. It can be observed that, under all tested PLMs, CSE-SFP consistently delivers the best performance. In particular, when employing Mistral\(_\text{7b}\), CSE-SFP surpasses PromptSTH, PromptSUM, and PromptEOL by 3.84\%, 6.32\%, and 9.02\%, respectively.

These results are encouraging, as the primary goal in designing CSE-SFP was to optimize efficiency, yet it realizes steady performance gains as well. This suggests that, compared to conducting two independent forward computations for data augmentation, CSE-SFP's two-stage template is more effective at leveraging contrastive learning to enhance the semantic space of PLMs. In Section~\ref{sec:analysis}, we will carry out a more rigorous analysis using mathematical tools to further investigate this phenomenon. Moreover, since CSE-SFP does not introduce any external components during the entire training or inference process, and relies solely on the model's intrinsic capabilities to achieve these improvements, this underscores the simplicity and effectiveness of our method.
\begin{table*}[!ht]
\caption{Spearman’s correlation scores for different methods on seven STS benchmarks under unsupervised settings.} 
\centering
    \begin{tabular}{ccccccccc}
    \toprule 
    \bf{Methods} & \bf{STS-12} & \bf{STS-13} & \bf{STS-14} & \bf{STS-15} & \bf{STS-16} & \bf{STS-B} & \bf{SICK-R} & \bf{Avg.} \\
    \midrule
    \multicolumn{9}{c}{\it{Implementation on $\rm{LLaMA2}_{\rm 7b}$}} \\
    PromptEOL & 70.35 & 86.29 & 78.75 & \bf 84.49 & 81.28 & 81.25 & 72.01 & 79.20 \\
    PromptSUM & 65.43 & 84.09 & 75.32 & 80.22 & 78.83 & 74.80 & 69.78 & 75.50 \\
    PromptSTH & 65.18 & 81.99 & 72.28 & 78.58 & 78.16 & 72.68 & 67.77 & 73.81 \\
    CSE-SFP & \bf 71.94 & \bf 86.79 & \bf 79.60 & 83.64 & \bf 81.78 & \bf 82.97 & \bf 74.14 & \bf 80.12 \\
    \midrule
    \multicolumn{9}{c}{\it{Implementation on $\rm{LLaMA3}_{\rm 8b}$}} \\
    PromptEOL & 68.63 & 86.17 & 78.39 & 84.47 & 81.40 & 81.25 & 73.08 & 79.06 \\
    PromptSUM & 62.59 & 83.00 & 75.57 & 81.56 & 77.94 & 76.75 & 71.75 & 75.59 \\
    PromptSTH & 63.69 & 80.72 & 74.66 & 80.57 & 79.30 & 76.25 & 69.99 & 75.03 \\
    CSE-SFP & \bf 70.27 & \bf 86.80 & \bf 79.56 & \bf 86.02 & \bf 82.24 & \bf 82.46 & \bf 75.02 & \bf 80.34 \\
    \midrule
    \multicolumn{9}{c}{\it{Implementation on $\rm{OPT}_{\rm 6.7b}$}} \\
    PromptEOL & \bf 68.85 & 83.28 & 75.51 & 83.56 & 81.24 & 79.52 & 69.56 & 77.36 \\
    PromptSUM & 67.98 & \bf 84.31 & 76.78 & 84.32 & 81.47 & 81.21 & 71.75 & 78.26 \\
    PromptSTH & 68.68 & 83.44 & 76.48 & 83.55 & \bf 82.58 & 80.31 & \bf 72.18 & 78.17 \\
    CSE-SFP & 67.83 & 84.11 & \bf 77.53 & \bf 84.41 & 82.35 & \bf 81.78 & 71.75 & \bf 78.54 \\
    \midrule
    \multicolumn{9}{c}{\it{Implementation on $\rm{Mistral}_{\rm 7b}$}} \\
    PromptEOL & 59.59 & 74.72 & 69.89 & 76.64 & 75.20 & 71.18 & 60.55 & 69.68 \\
    PromptSUM & 56.81 & 78.59 & 72.76 & 78.10 & 74.68 & 74.78 & 70.92 & 72.38 \\
    PromptSTH & 67.44 & 80.81 & 73.09 & 79.71 & 80.99 & 74.78 & 67.19 & 74.86 \\
    CSE-SFP & \bf 68.07 & \bf 85.62 & \bf 78.77 & \bf 84.10 & \bf 83.05 & \bf 79.49 & \bf 71.77 & \bf 78.70 \\
    \bottomrule
    \end{tabular}
\label{tab:sts_results}
\end{table*}

Additionally, when LLaMA3\(_\text{8b}\) serves as the backbone, CSE-SFP attains an average Spearman correlation score of 80.34, significantly higher than the 76.25 obtained by SimCSE-BERT\(_{\text{base}}\) \cite{SimCSE-EMNLP-2021}. This result reflects the advantages of more powerful and well-trained LLMs in embedding derivation. Currently, an increasing number of generative PLMs have exceeded the 6-8 billion parameter range, reaching scales of tens or even hundreds of billions \cite{Chinchilla-2022, MTNLG-2022, OPT-2022, PaLM-2023, LLaMA2-2023, LLaMA3-2024}. The introduction of CSE-SFP opens new possibilities for combining these advanced LLMs with unsupervised text representation learning.

\subsection{Performance on Information Retrieval Tasks}
\label{sec:ir_performance}

Beyond STS benchmarks, we further evaluate model performance on eight IR tasks via the MTEB leaderboard. Following the same testing procedure as described in subsection~\ref{sec:sts_performance}, we directly load the model checkpoints fine-tuned on the Wiki-1M dataset through contrastive learning, without performing any additional parameter updates or structural modifications specific to each task. In fact, the checkpoints utilized in these two subsections are completely identical. This zero-shot evaluation setup will maximally reflect the transferability of our method.

Given that tasks on the MTEB leaderboard are typically large in scale and require substantial testing time \cite{E5Mistral-2023}, we opt to conduct experiments using Mistral\(_\text{7b}\) and LLaMA3\(_\text{8b}\), which are among the most popular PLMs currently available. Table~\ref{tab:ir_results} summarizes the results, where "PLM-Raw" refers to the original Mistral\(_\text{7b}\) and LLaMA3\(_\text{8b}\) models. As shown, without the enhancement of contrastive learning, even extensively pre-trained LLMs like Mistral and LLaMA3 struggle with complex IR tasks. Across all eight benchmarks, the "PLM-Raw" scores are consistently below 8\% for each task.

With the aid of prompt engineering and contrastive learning, the output vectors from PromptEOL, PromptSUM, and PromptSTH exhibit significant improvements over the raw embeddings of the PLMs. More impressively, our proposed CSE-SFP largely outperforms these state-of-the-art methods, achieving the best results in all eight tasks. Specifically, when Mistral\(_\text{7b}\) serves as the backbone, CSE-SFP surpasses the baselines by more than 10 percentage points in half of the eight tasks: LEMBSummScreenFD, SciFact, MedicalQA, and LegalSumm. Similarly, when leveraging LLaMA3\(_\text{8b}\), CSE-SFP also demonstrates outstanding performance. For example, on the SpartQA benchmark, CSE-SFP’s score exceeds those of the other methods by over tenfold. 

Considering that the training sets, loss functions, and QLoRA configurations for PromptEOL, PromptSUM, PromptSTH, and CSE-SFP remain all the same throughout our experiments, this provides compelling evidence for the superiority of CSE-SFP's representation derivation strategy. Moreover, given CSE-SFP's robust performance across multiple tasks and its strong adaptability, it may offer additional benefits in scenarios with scarce labeled data, as downstream neural networks can harness the embeddings produced by CSE-SFP as initial features to further enhance performance.
\begin{table*}[!ht]
\caption{Performance of different models on eight IR benchmarks. The reported values correspond to the primary evaluation metrics for each task, scaled to a percentage format by multiplying by 100.} 
\centering
\resizebox{1.0\linewidth}{!}{
    \begin{tabular}{ccccccccc}
    \toprule 
    \bf Methods & \bf LEMBSummScreenFD & \bf ARCChallenge & \bf SciFact & \bf SpartQA & \bf MedicalQA & \bf NFCorpus & \bf LegalSumm & \bf LegalBenchCorporateLobbying \\
    \midrule
    \multicolumn{9}{c}{\it{Implementation on $\rm{Mistral}_{\rm 7b}$}} \\
    Mistral-Raw & 5.72 & 1.61 & 1.64 & 0.11 & 7.35 & 2.57 & 10.11 & 3.70 \\
    PromptEOL & 28.88 & 5.29 & 45.25 & 1.16 & 30.39 & 15.44 & 58.32 & 88.81 \\
    PromptSUM & 22.06 & 7.00 & 32.78 & 0.22 & 33.37 & 17.93 & 56.12 & 75.09 \\
    PromptSTH & 19.20 & 7.51 & 42.94 & 7.55 & 26.77 & 21.49 & 51.48 & 70.27 \\
    CSE-SFP & \bf 42.05 & \bf 13.58 & \bf 60.31 & \bf 11.22 & \bf 50.55 & \bf 25.56 & \bf 69.61 & \bf 89.25 \\
    \midrule
    \multicolumn{9}{c}{\it{Implementation on $\rm{LLaMA3}_{\rm 8b}$}} \\
    LLaMA3-Raw & 6.33 & 2.90 & 3.04 & 0.09 & 7.33 & 3.41 & 6.63 & 3.87 \\
    PromptEOL & 25.47 & 15.73 & 55.21 & 0.45 & 54.31 & 27.67 & 63.31 & 89.69 \\
    PromptSUM & 46.82 & 16.27 & 64.63 & 0.04 & 60.04 & 29.68 & 63.44 & 90.48 \\
    PromptSTH & 39.89 & 13.46 & 62.22 & 0.57 & 57.03 & 28.81 & 57.32 & 88.39 \\
    CSE-SFP & \bf 47.16 & \bf 17.11 & \bf 64.81 & \bf 8.85 & \bf 61.99 & \bf 32.41 & \bf 68.65 & \bf 91.79 \\
    \bottomrule
    \end{tabular}%
}
\label{tab:ir_results}
\end{table*}

\subsection{Computational Cost Comparison}
\label{sec:comparison}

As demonstrated above, CSE-SFP excels in both semantic capture and text matching. In this subsection, we further confirm that CSE-SFP not only generates high-quality sentence representations but is more computationally efficient as well.

We compare the GPU memory consumption and training time of CSE-SFP with those of mainstream contrastive learning methods using four RTX 4090 GPUs. For a fair comparison, we uniformly set the number of epochs to 1, the batch size to 256, and the truncation length to 32. All other experimental settings are consistent with the descriptions in subsection~\ref{sec:implementation}.

The results, presented in Table~\ref{tab:cost}, indicate that CSE-SFP outperforms conventional contrastive learning approaches in both time and memory efficiency. For instance, when employing LLaMA3\(_\text{8b}\) as the PLM, even with parameter-efficient fine-tuning techniques, PromptEOL still takes 280.84 minutes to complete training and consumes 93.33 GB (95,568 MB) of GPU memory. In contrast, CSE-SFP accelerates the training speed by 43\% and frees up approximately 8 GB (7,905 MB) of memory usage. This highlights that simplifying the two forward computations required for constructing positive sample pairs into a single pass significantly reduces the computational overhead of contrastive learning. As model sizes and dataset scales continue to increase, the advantages of CSE-SFP will become even more pronounced.

Furthermore, combining the experimental results from subsections~\ref{sec:sts_performance} and~\ref{sec:ir_performance}, it is clear that CSE-SFP not just optimizes efficiency, it can also deliver superior performance, making it a viable option for deployment in a wide range of applications.
\begin{table}[htbp]
\caption{Training time and computational resource consumption for different text representation methods during parameter updates.} 
\centering
\resizebox{1.0\linewidth}{!}{
\begin{tabular}{c|c|c|c}
    \toprule
    \bf PLMs & \bf Methods & \bf Training Time & \bf Memory Usage \\
    \midrule
    \multirow{4}{*}{LLaMA2$_{\rm 7b}$} 
    & PromptEOL & 265.48 min & 79.63 GB \\
    & PromptSUM & 265.05 min & 79.63 GB \\
    & PromptSTH & 265.63 min & 79.68 GB \\
    & CSE-SFP & \textbf{169.30} min & \textbf{71.70} GB \\
    \midrule
    \multirow{4}{*}{LLaMA3$_{\rm 8b}$} 
    & PromptEOL & 280.84 min & 93.33 GB \\
    & PromptSUM & 280.65 min & 93.33 GB \\
    & PromptSTH & 242.19 min & 91.47 GB \\
    & CSE-SFP & \textbf{159.27} min & \textbf{85.61} GB \\
    \midrule
    \multirow{4}{*}{OPT$_{\rm 6.7b}$} 
    & PromptEOL & 234.07 min & 78.96 GB \\
    & PromptSUM & 234.22 min & 78.96 GB \\
    & PromptSTH & 199.31 min & 76.80 GB \\
    & CSE-SFP & \textbf{129.42} min & \textbf{72.00} GB \\
    \midrule
    \multirow{4}{*}{Mistral$_{\rm 7b}$} 
    & PromptEOL & 292.92 min & 85.82 GB \\
    & PromptSUM & 292.83 min & 85.82 GB \\
    & PromptSTH & 292.88 min & 85.84 GB \\
    & CSE-SFP & \textbf{189.68} min & \textbf{80.29} GB \\
    \bottomrule
\end{tabular}
}
\label{tab:cost}
\end{table}

\section{Analysis}
\label{sec:analysis}

This section analyzes the reasons behind the effectiveness of CSE-SFP. First, in subsection~\ref{sec:align_uniform}, we assess whether the sentence representations derived from CSE-SFP exhibit superior semantic distinction by utilizing two critical metrics that reflect the distributional characteristics of embeddings: alignment and uniformity. Specifically, we also propose two additional ratio-based metrics to  facilitate a more comprehensive evaluation. Then, in subsection~\ref{sec:aniso_smooth}, we explore the alleviating effects of CSE-SFP on anisotropy and over-smoothing issues by examining the singular values of the word vector matrix and the similarity between token embeddings.

\subsection{Alignment and Uniformity}
\label{sec:align_uniform}
\begin{table*}[ht]
\caption{Performance of various unsupervised sentence embedding derivation methods on the STS-B and SICK-R test sets. Higher values in the Spearman column are better, while lower values in the Alignment, Uniformity, Ratio 1, and Ratio 2 columns are preferred.}
\centering
\begin{tabular}{cccccc}
\toprule
\bf Methods & \bf Spearman & \bf Alignment & \bf Uniformity & \bf Ratio 1 & \bf Ratio 2\\
\midrule
\multicolumn{6}{c}{\it{Calculation based on the STS-B test set}} \\
PromptSTH & \underline{74.78} & 0.3927 & -3.3815 & \underline{0.2274} & \underline{0.2552} \\
PromptSUM & 74.78 & 0.4319 & \bf -3.5248 & 0.2397 & 0.2666 \\
PromptEOL & 71.18 & 0.5185 & -3.5101 & 0.2897 & 0.3359 \\
CSE-SFP & \bf 79.49 & \bf 0.2326 & -3.1289 & \bf 0.1429 & \bf 0.1524  \\
\midrule
\multicolumn{6}{c}{\it{Calculation based on the SICK-R test set}} \\
PromptSTH & 67.19 & 0.3968 & -2.8446 & 0.2614 & 0.3168 \\
PromptSUM & \underline{70.92} & 0.3973 & -3.0802 & \underline{0.2337} & \underline{0.2900} \\
PromptEOL & 60.55 & 0.4570 & \bf -3.1914 & 0.2612 & 0.3339 \\
CSE-SFP & \bf 71.77 & \bf 0.2275 & -2.5684 & \bf 0.1819 & \bf 0.1883 \\
\bottomrule
\end{tabular}
\label{tab:ratio}
\end{table*}

In representation learning, alignment and uniformity \cite{AlignUniform-ICML-2020} are widely adopted to assess the properties of a model's semantic space. Alignment measures how tightly the embeddings of positive sample pairs are distributed. As shown in Equation~\ref{eq:align}, a lower alignment value indicates that embeddings for semantically similar texts are closer together, thus enabling more effective reflection through standard distance metrics.
\begin{equation}
    \label{eq:align}
    \ell_\text{align} \triangleq \mathbb{E}_{(x, x^+) \sim p_{\text{data}}} \|f(x) - f(x^+)\|^2
\end{equation}

In contrast, uniformity evaluates the overall evenness of the embedding space by computing the distances between unrelated samples. Owing to the negative sign in Equation~\ref{eq:uniform}, uniformity also benefits from a smaller value, as it suggests that sentence vectors of different types are more evenly distributed on the high-dimensional hypersphere and do not cluster too densely in specific regions. However, in real-world scenarios, due to the lack of annotated information, there may occasionally be semantic correlations between "unrelated" pairs $(x, y)$ (i.e., false negatives), which could introduce noise into the results.
\begin{equation}
    \label{eq:uniform}
    \ell_\text{uniform} \triangleq \log \mathbb{E}_{x, y \sim p_{\text{data}}} e^{-2 \| f(x) - f(y) \|^2}
\end{equation}

Mathematically, alignment and uniformity are inherently competing objectives. Over-optimizing uniformity can potentially degrade alignment, and vice versa. Consequently, when these metrics are used as loss functions, a weighted mechanism is often employed to strike a balance. Nevertheless, many researchers in contrastive learning treat alignment and uniformity as independent criteria and analyze them separately \cite{PT-BERT-ACL-2022, RankCSE-ACL-2023}. It should be noted that the similarity or distance between two pieces of text may not hold much substantive meaning; what truly matters is the ordinal relationship between these scores, which is why Spearman's rank correlation is regarded the core metric in STS tasks \cite{Pcc-tuning-EMNLP-2024}.

Thus, we argue that alignment and uniformity should be considered in a more integrated manner. Techniques that perform weakly in one of these metrics might still yield a more favorable semantic space by significantly improving the other. For example, the SOTA strategy for BERT-based sentence representations, CoT-BERT \cite{CoT-BERT-ICANN-2024}, found that by introducing additional reference terms into the InfoNCE loss, although the alignment of the embedding space decreased, the uniformity and downstream task performance consistently improved.

Furthermore, combining alignment and uniformity into a more comprehensive metric offers a potential advantage: it can serve as a decisive tie-breaker in many "draw" scenarios. A "draw" occurs when comparing two semantic encoders, A and B, where A excels in alignment and B excels in uniformity. In such cases, it is typically hard to determine which embedding distribution is superior. By introducing a more holistic metric, we can resolve this ambiguity and identify the optimum strategy when multiple Pareto-optimal solutions exist.

Since both alignment and uniformity are preferred to have lower values, we seek to design a unified metric that follows this same pattern. Based on this idea, we place the distance computation for positive sample pairs (emphasized in alignment) in the numerator, and the distance calculation between unrelated text embeddings (emphasized in uniformity) in the denominator. To avoid discrepancies in the numerical ranges due to the distinct expressions of alignment and uniformity, we adjust their respective formulas and derive the following two ratio-based metrics:
\begin{equation}
\begin{aligned}
\text{Ratio 1} &= \frac{\mathbb{E}_{(x, x^+) \sim p_{\text{data}}} \| f(x) - f(x^+) \|^2}{\mathbb{E}_{x, y \stackrel{i.i.d.}{\sim} p_{\text{data}}} \| f(x) - f(y) \|^2} \\
\text{Ratio 2} &= \frac{\log \mathbb{E}_{(x, x^+) \sim p_{\text{data}}} e^{2 \| f(x) - f(x^+) \|^2}}{\log \mathbb{E}_{x, y \stackrel{i.i.d.}{\sim} p_{\text{data}}} e^{2 \| f(x) - f(y) \|^2}}
\end{aligned}
\end{equation}

Ratio 1 and Ratio 2 are conceptually similar, but differ in their computational approach, which corresponds to the original formulas for alignment and uniformity, respectively. Lower values for both ratios indicate that the model tightly encodes positive pairs while maximizing the separation between negative pairs, thereby demonstrating superior semantic differentiation. Compared to separately measuring alignment or uniformity, these ratios provide a more reasonable and comprehensive evaluation. Specifically, suppose that the alignment of a PLM is negatively affected during the usage of a given method (i.e., the distance between semantically similar vectors increases). However, as long as the distances among unrelated embeddings increase even more, the model’s overall semantic space will still improve. This enhancement, driven by sacrificing one metric to substantially boost the other, can also be captured by our ratio metrics, as both Ratio 1 and Ratio 2 will decrease in such cases.

Using Mistral\(_\text{7b}\) as the backbone, we compute various metrics for different sentence representation methods on the STS-B and SICK-R test sets, with the results presented in Table~\ref{tab:ratio}. It can be observed that, compared to PromptEOL, PromptSTH, and PromptSUM, although CSE-SFP does not lead in uniformity, it far surpasses the other methods in alignment, ultimately attaining superior scores in both Ratio 1 and Ratio 2. This proves that CSE-SFP produces a more favorable embedding distribution. Moreover, there is a strong correlation between the ratios and Spearman's correlation coefficient. Methods that rank in the top two for Spearman’s correlation also perform similarly in Ratio 1 and Ratio 2, further confirming that CSE-SFP optimizes the PLM semantic space more effectively than traditional contrastive learning approaches.
\begin{table*}[ht]
\caption{Average token embedding similarity, condition number, and singular value entropy for various methods on the STS-B test set. Lower values for Token-wise Similarity and Condition Number are preferred, while higher values for Singular Values Entropy are desirable.}
\centering
\begin{tabular}{cccc}
\toprule
\bf Methods & \bf Token-wise Similarity & \bf Condition Number & \bf Singular Values Entropy\\
\midrule
Mistral-Raw & 0.4203 & 7.2334 & 1.6870 \\
PromptSTH & 0.3909 & 6.6636 & 1.7574 \\
PromptSUM & 0.3847 & 6.5490 & 1.7720 \\
PromptEOL & 0.4038 & 6.8993 & 1.7272 \\
CSE-SFP & \bf 0.3622 & \bf 6.2164 & \bf 1.8235 \\
\bottomrule
\end{tabular}
\label{tab:math}
\end{table*}

\subsection{Over-smoothing and Anisotropy Issues}
\label{sec:aniso_smooth}

The issues of anisotropy and over-smoothing in PLMs pose significant challenges to sentence representation research. Anisotropy \cite{Anisotropy-EMNLP-2019} can be interpreted as the phenomenon in which parameter updates of neural networks are influenced by factors such as word frequency \cite{BERT-flow-EMNLP-2020}, capitalization \cite{CARDS-SIGIR-2022}, punctuation, and subword tokenization \cite{PromptBERT-EMNLP-2022}, causing the output embeddings to exhibit clear biases and concentrate in a narrow zone of high-dimensional space. Over-smoothing \cite{Over-smoothing-2022}, on the other hand, refers to the situation where different parts of an input sentence, when mapped to token embeddings, show excessive similarity. In other words, the model loses its ability to distinguish between words during encoding.

Both phenomena negatively impact sentence representation quality. This is because, whether through prompt engineering or pooling, existing text representation methods essentially rely on token embeddings to approximate sentence embeddings. Therefore, any biases or information loss in token embeddings degrade the model's ability to accurately represent the entire sentence.

We can quantify the degree of over-smoothing and anisotropy within a model via mathematical tools. Given an input sentence \(T = [t_1, t_2, \ldots, t_n]\), the PLM outputs a word embedding matrix \(X = \{x_1, x_2, \ldots, x_n\}\), where each \(x_i\) is a vector of the hidden layer's dimension. Obviously, the higher the token-wise cosine similarity in \(X\), the more severe the over-smoothing \cite{SSCL-ACL-2023}:
\begin{equation}
    \text{TokSim} = \frac{1}{n(n-1)} \sum_{i \neq j} \frac{x_i^T x_j}{\|x_i\|_2 \|x_j\|_2}
\end{equation}

Likewise, we can analyze the singular value distribution of \( X \) to assess the efficacy of contrastive learning in alleviating anisotropy. Here, we leverage the condition number and entropy of the singular values as indicators. The condition number is defined as the ratio between the largest and smallest singular values. A smaller condition number typically signifies a more uniform distribution of singular values:
\begin{equation}
    \kappa = \frac{\sigma_{\max}}{\sigma_{\min}}
\end{equation}

Entropy can also describe the evenness of the singular values in the word embedding matrix \( X \). To compute entropy, we first normalize the singular values \(\sigma_i\) into a probability distribution, and then apply the following formula:
\begin{equation}
\begin{aligned}
p_i &= \frac{\sigma_i^2}{\sum_{j=1}^m \sigma_j^2} \\
\text{Entropy} &= -\sum_{i=1}^m p_i \log(p_i)
\end{aligned}
\end{equation}

Higher entropy indicates that more dimensions contribute to the valid information of the matrix. In the context of text representation, we aim for the model to capture features from multiple aspects, thereby mitigating the impact of erroneous priors and enhancing the robustness of embeddings. Previous studies, such as OssCSE \cite{OssCSE-EMNLP-2023}, SNCSE \cite{SNCSE-ICIC-2023} and PT-BERT \cite{PT-BERT-ACL-2022}, have shown that PLMs fine-tuned on unsupervised corpora tend to learn incorrect surface structure biases. Therefore, if the token embedding matrix has a high condition number and low entropy, it suggests that the PLM primarily relies on a few dominant components during encoding, which could limit its ability to discern fine-grained semantic distinctions.

Leveraging Mistral\(_\text{7b}\) as the PLM, we compute the average token similarity, condition number, and singular value entropy for various sentence representation methods on the STS-B test set. The results are recorded in Table~\ref{tab:math}. As previously observed in Table~\ref{tab:ir_results}, Mistral's raw embeddings perform poorly across all three metrics. PromptSTH, PromptSUM, and PromptEOL show notable improvements over the baseline. In comparison, CSE-SFP further strengthens the effects of contrastive learning through more efficacious positive sample construction, achieving the best results in all metrics. Therefore, we can conclude that CSE-SFP’s ability to generate high-quality sentence embeddings is partly attributed to its mitigation of over-smoothing and anisotropy issues within the PLM semantic space.

\section{Conclusion}

This paper presents CSE-SFP, an unsupervised sentence representation framework that realizes effective contrastive learning with only a single forward pass. We thoroughly validate CSE-SFP's superiority in both performance and efficiency across multiple PLMs and various STS and IR tasks. Additionally, we propose two novel ratio-based metrics built upon alignment and uniformity, which offer a more comprehensive evaluation of models' semantic space. Furthermore, we also conduct an in-depth analysis to uncover the underlying factors that contribute to the success of CSE-SFP.


\bibliographystyle{ACM-Reference-Format}
\balance
\bibliography{sample-base}


\end{document}